\documentclass[letterpaper, 10 pt, conference]{ieeeconf}  %

\IEEEoverridecommandlockouts
\usepackage{cite}
\usepackage{amsmath,amssymb,amsfonts}
\usepackage{algorithmic}
\usepackage{graphicx}
\usepackage{textcomp}
\usepackage{xcolor}
\usepackage{balance}
\usepackage{comment}
\usepackage[ruled,vlined]{algorithm2e}
\usepackage{subcaption}
\usepackage{tensor}

\newcommand{\ie}{\emph{i.e.},}

\usepackage{hyperref}
\usepackage{float}

\def\secref#1{Sec.~\ref{#1}}
\def\figref#1{Fig.~\ref{#1}}
\def\tabref#1{Table~\ref{#1}}
\def\eqref#1{Eq.~(\ref{#1})}

\DeclareMathOperator*{\argmin}{argmin}

\def\BibTeX{{\rm B\kern-.05em{\sc i\kern-.025em b}\kern-.08em
    T\kern-.1667em\lower.7ex\hbox{E}\kern-.125emX}}
\begin{document}

\title{\LARGE \bf LoGG3D-Net: Locally Guided Global Descriptor Learning \\for 3D Place Recognition %
}

\author{Kavisha Vidanapathirana$^{1,2}$, Milad Ramezani$^{1}$, Peyman Moghadam$^{1,2}$, \\ Sridha Sridharan$^{2}$, Clinton Fookes$^{2}$   
\thanks{
$^1$ Kavisha Vidanapathirana, Milad Ramezani and Peyman Moghadam are with the Robotics and Autonomous Systems Group, DATA61, CSIRO, Brisbane, QLD 4069, Australia. 
E-mails: {\tt\footnotesize \emph{
firstname.lastname
}@data61.csiro.au}}
\thanks{
$^{2}$ Kavisha Vidanapathirana, Peyman Moghadam, Sridha Sridharan, Clinton Fookes are with the SAIVT research programme in the School of Electrical Engineering and Robotics, Queensland University of Technology (QUT), Brisbane, Australia.
E-mails: {\tt\footnotesize \emph\{kavisha.vidanapathirana, peyman.moghadam, s.sridharan, c.fookes\}@qut.edu.au}
}
}

\maketitle

\begin{abstract}
Retrieval-based place recognition is an efficient and effective solution for re-localization within a pre-built map, or global data association for Simultaneous Localization and Mapping (SLAM). The accuracy of such an approach is heavily dependent on the quality of the extracted scene-level representation. While end-to-end solutions - which learn a global descriptor from input point clouds - have demonstrated promising results, such approaches are limited in their ability to enforce desirable properties at the local feature level. In this paper, we introduce a local consistency loss to guide the network towards learning local features which are consistent across revisits, hence leading to more repeatable global descriptors resulting in an overall improvement in 3D place recognition performance. We formulate our approach in an end-to-end trainable architecture called \textit{LoGG3D-Net}.
Experiments on two large-scale public benchmarks (KITTI and MulRan) show that our method achieves mean $F1_{max}$ scores of $0.939$ and $0.968$ on KITTI and MulRan respectively, achieving state-of-the-art performance while operating in near real-time. The open-source implementation is available at: \href{https://github.com/csiro-robotics/LoGG3D-Net}{https://github.com/csiro-robotics/LoGG3D-Net}.

\end{abstract}

\section{Introduction}
\label{sec:intro}
Despite considerable progress in the field of 3D point cloud perception for robotics and self-driving cars, existing methods for data-association remain fragile and limited in applicability, especially in large-scale outdoor scenes. 
Accurate data-association is vital for enabling long-term autonomy, as autonomous agents need to construct and maintain an accurate representation of the environment they operate in. 
Recognizing previously visited places provides global constraints to restrict cumulative errors within Simultaneous Localization and Mapping (SLAM) systems \cite{Park2018,park2021elasticity}. %
This is known as the Place Recognition (PR) task. 

Compared to visual place recognition \cite{Lowry2016}, the use of 3D point clouds extracted from LiDAR sensors benefits from its inherent invariance to view-point and illumination. However extracting useful information from the point cloud representation remains a challenge due to higher sparsity and a complex and variable distribution of points.
In this paper, we consider the task of place recognition on 3D point clouds.  

While there have been many handcrafted approaches for extracting useful information for the task of place recognition \cite{He2016, Kim2018, Salti2014}, discriminative learning-based  approaches have demonstrated competitive performance in terms of both accuracy and efficiency~\cite{Uy2018, Komorowski_2021_WACV}. %

\begin{figure}[t!]
\centering
\includegraphics[width=0.48\textwidth]{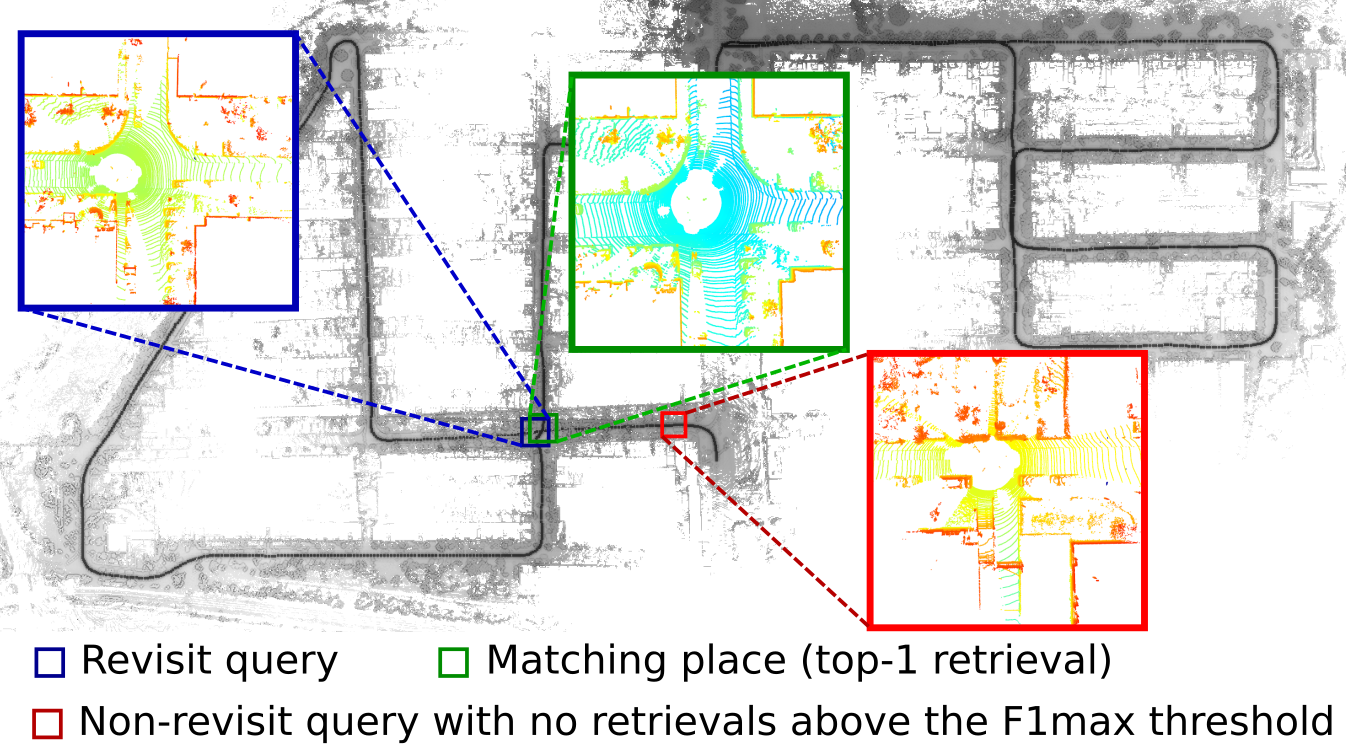}
\caption{\textit{LoGG3D-Net} retrieval results for challenging orthogonal revisits and non-revisit queries on KITTI sequence 08.}
\label{fig:front_page}
\end{figure}

We propose \textit{LoGG3D-Net}, a novel end-to-end 3D place recognition method for LiDAR point clouds. \figref{fig:front_page} shows an example of retrieval results on KITTI sequence 08 which demonstrates some challenging orthogonal revisits. In contrast to state-of-the-art end-to-end methods, which rely solely on global descriptor learning, we propose to jointly optimize local and scene-level embeddings. We achieve this by introducing a local consistency loss that takes a pair of LiDAR point clouds and maximizes the similarity of their corresponding points’ features and minimizes the similarity of all non-corresponding points’ features.%

We further introduce the use of second-order pooling followed by differentiable Eigen-value power normalization to aggregate local features and generate a global descriptor in an end-to-end setting. We note that all current learning-based methods use NetVLAD\cite{Arandjelovic2016} or other first-order aggregation methods \cite{gem19} to compute the global descriptor, and higher-order aggregation methods have not been explored in an end-to-end setting for LiDAR-based place recognition. We evaluate the accuracy and robustness of \textit{LoGG3D-Net} on 6 KITTI sequences and 5 sequences of the MulRan dataset. Note that these datasets are collected using different LiDAR sensors (Velodyne HDL-64E in KITTI, Ouster OS1-64 in MulRan) and different countries (Germany, Korea). Our main contributions are summarized as follows: 
\begin{itemize}
    \item We introduce a local consistency loss that can be used in an end-to-end global descriptor learning setting to enforce consistency of the local embeddings extracted from point clouds of the same location. We demonstrate how enforcing this property in the local features contributes towards better performance of the global descriptor. 
    \item We introduce the use of second-order pooling with differentiable Eigen-value power normalization in an end-to-end setting for LiDAR-based  place recognition.
    \item Using 2 large scale public datasets, we demonstrate the superiority of our method in an end-to-end setting while operating in near real-time.
\end{itemize}

\begin{figure*}[ht]
    \centering
    \includegraphics[width=0.8\textwidth]{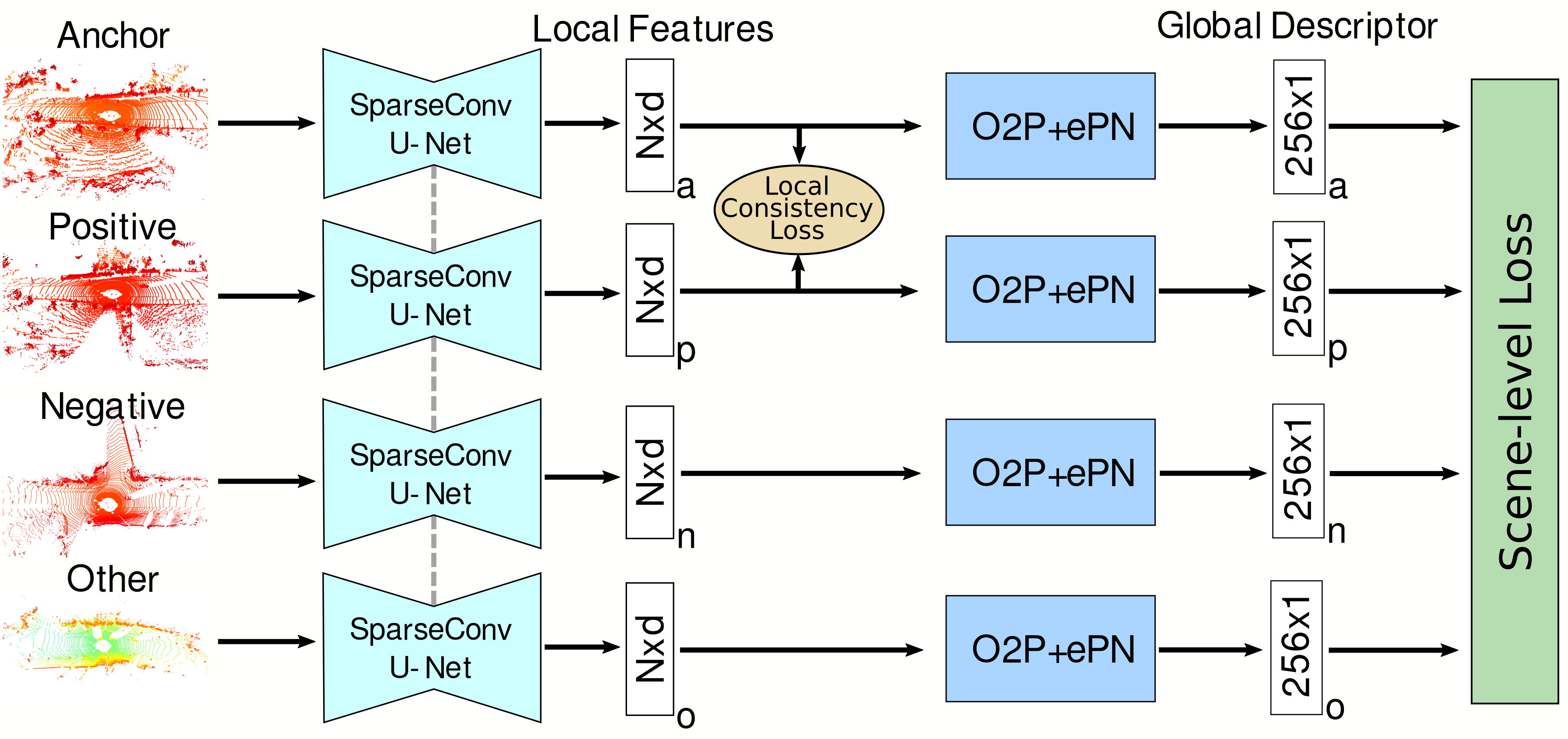}
    \caption{An overview of the proposed \textit{LoGG3D-Net} method. A tuple of point clouds are fed into a SparseConv U-Net. The local consistency loss is applied to the anchor and positive point clouds to enforce consistency of local feature embeddings. Local features are aggregated using second-order pooling (O2P) followed by Eigen-value Power Normalization (ePN) to obtain a global descriptor. The final loss is the combination of the local consistency loss with the scene-level loss. }
    \label{fig:bl_pipeline}
\end{figure*}

The remainder of this paper is organized as follows. In~\secref{sec:rel_work} we discuss related work.~\secref{sec:method} explains our proposed method. The experimental setup is outlined in~\secref{sec:experiments}.~\secref{sec:results} discusses the experimental results and concluding remarks are provided in~\secref{sec:conclusion}.

\section{Related Work}
\label{sec:rel_work}

Solutions for 3D LiDAR-based place recognition can be loosely classified under local feature matching based methods and retrieval-based methods. Local feature matching based methods rely on accurate global pose estimates to limit the search space of feature correspondences and hence do not scale well for the task of re-localization in large-scale environments. Majority of local feature matching based approaches operate by detecting keypoints and describing their local neighborhood \cite{Salti2014,Guo2019}. Recently features extracted from larger local regions such as point segments have shown superior performance \cite{segmap2020,esm19}.  

Retrieval-based methods are inexpensive due to simple matching of global descriptors and therefore scale well to large-scale scenarios. They operate by encoding each point cloud to a single vector representation that can be used for querying a database of previously visited places. Global descriptors can be categorized as handcrafted \cite{He2016, Kim2018}, hybrid \cite{vidanapathirana2020locus}, and end-to-end learning-based \cite{Uy2018}. Handcrafted methods have the benefit of not needing re-training to adapt to different environments and sensor-types. Prominent methods such as  ScanContext \cite{Kim2018} have demonstrated reliable performance in various scenarios. However the discriminative power of such methods remains limited.

Hybrid methods aim to combine principled mathematical models with data-driven models to benefit from the advantages of both \cite{9523403}. Locus \cite{vidanapathirana2020locus} first demonstrated such an approach for lidar-based PR by mathematically modeling the topological relationships and temporal consistencies of point segments, while structural appearance of the segments were encoded using a data-driven 3D-CNN. Locus achieved state-of-the-art on the KITTI dataset, but struggles to adapt to environments where extracted point segments are structurally different to its training data (eg. unstructured environments). While (in general) hybrid methods show promise in scenarios with inadequate training data \cite{9523403}, the universal function approximation property of neural networks \cite{hornik1989multilayer} implies that end-to-end methods have the potential to perform better if trained properly. This paper presents an improved method of training of end-to-end models for lidar-based PR.

End-to-end methods formulate the learning of global descriptors using a contrastive approach to obtain discriminative scene representations. 
PointNetVLAD\cite{Uy2018} pioneered the use of an end-to-end trainable global descriptor for 3D point cloud place recognition. PointNetVLAD extracts local features from PointNet~\cite{Qi_2017_CVPR} and employs the NetVLAD aggregator~\cite{Arandjelovic2016} to form a global descriptor of the scene. To address the limited descriptive power of the PointNet backbone, other approaches such as LPD-Net \cite{lpdnet_new} have been proposed. More recently MinkLoc3D\cite{Komorowski_2021_WACV} proposed an efficient architecture outperforming its end-to-end predecessors. MinkLoc3D utilized sparse convolutions which have been demonstrated to be effective at capturing useful point-level features. We also benefit from the use of sparse convolutions. 

We note that all end-to-end methods rely solely on supervisory signals applied on the global descriptors. Additionally, all end-to-end methods are currently utilizing first-order aggregation methods in the formulation of the global descriptors. In many visual recognition tasks, higher-order aggregation methods have demonstrated superior performance \cite{7439823,9521687, Li2017}. Higher-order aggregation has previously been applied for 3D place recognition \cite{vidanapathirana2020locus}, but not in an end-to-end trainable architecture. In this work, we address the above limitations by introducing an additional training signal to the local features, and using differentiable second-order pooling for global descriptor generation respectively. 

\section{Proposed Method}
\label{sec:method}

The overall architecture of our \textit{LoGG3D-Net} is presented in~\figref{fig:bl_pipeline}. Given a raw point cloud input, a sparse convolution-based U-Net (SparseConv U-Net) is used to embed each point into a high dimensional feature space. 
The ‘Local Consistency Loss’ acts on a pair of point clouds from nearby locations (with considerable overlap of points) to maximize the similarity of corresponding points’ features in the embedding space.
Next, local features are aggregated using second-order pooling followed by differentiable Eigen-value power normalization to form a global scene-level descriptor. A quadruplet loss is used to train our scene-level global descriptors. Finally, we combine our local consistency loss with our scene-level loss to optimize our network.

\subsection{Problem Formulation}
\label{section:formulation}
The task of point cloud based retrieval for place recognition is generally formulated as follows. Given a point cloud $ \mathcal{P} \in \mathbb{R}^{N \times 4} $ representing varying number of $N$ points with associated $x, y, z$ and intensity, a mapping function $ \Phi : \mathcal{P} \rightarrow g \in \mathbb{R}^{d'} $ is developed that represents the point cloud with a fixed-size global descriptor $ g $. End-to-end learning of global descriptors is formulated as follows:

\textbf{Problem: }
Given a training set $ \{\mathcal{S}_i\} = \{ (\mathcal{P}_i,\mathtt{x}_i) \}  $ consisting of pairs of point cloud $ \mathcal{P}_i $ with associated geo-location $ \mathtt{x}_i $, find the parameters $ \theta $ of the mapping function $ \Phi_{\theta}  : \mathcal{P} \rightarrow g \in \mathbb{R}^{d'} $ , such that for any subset of samples $\{\mathcal{S}_a, \mathcal{S}_i, \mathcal{S}_j\}$,
\begin{equation}\label{eqn:problem_1}
  \mathcal{D}(\mathtt{x_a},\mathtt{x_i}) \leq  \mathcal{D}(\mathtt{x_a},\mathtt{x_j}) \implies \lVert g_a - g_i \rVert \leq  \lVert g_a - g_j \rVert,  
\end{equation}
where, $ \mathcal{D} $ represents geometric distance and $\lVert . \rVert$ represents a distance in the feature space (typically $ {L}_2$). 

Learning the parameters $ \theta $  that address the above problem is generally performed using a metric learning setting by applying a loss function $ \mathcal{L}_g $ that acts on the global descriptors extracted from a tuple $ \mathcal{T} $ of training samples. We use the subscript `$g$' in   $ \mathcal{L}_g $ to highlight that the loss is typically only applied on the global descriptors, \ie{} at the scene-level.

\subsection{Our Approach}
\label{section:approach}

We note that generally $ \Phi $ can be decomposed into two functions such that,
\begin{equation}\label{eqn:decomposition}
 \Phi \equiv \varphi \circ \phi ,
\end{equation}
where, $ \phi : \mathcal{P} \rightarrow \{f\} $ extracts local features $ f \in \mathbb{R}^{d} $, and $ \varphi : \{f\} \rightarrow g $ aggregates the local features into a single global descriptor $ g $. Under this setting, optimizing $\theta$ with a training signal applied only to the global descriptor seems limited, \ie{} there are desirable properties of the intermediate representations $ \{f\} $ (local features) that cannot be fully enforced by $ \mathcal{L}_g $.

\textbf{Hypothesis: }
Given two point clouds from nearby locations (\ie{} with considerable overlap of points), enforcing consistency of local features of corresponding points will result in a more repeatable global descriptors after aggregation. %

We define `corresponding points' as two points from different point clouds that are nearby when represented with respect to a global coordinate frame. `Consistency of local features' implies that local features of corresponding points should be nearby in the embedding space. Towards testing the above hypothesis, we introduce an additional training signal to enforce the features  $ \{f\} $ of corresponding points to be consistent in the embedding space. 

\subsection{Local Descriptor}
\label{section:local_feat}

For the local feature extractor $ \phi $, we use a Sparse U-Net style backbone~\cite{tang2020searching} which performs sparse point-voxel convolution. The backbone consists of two branches: a voxel-based branch and a point-based branch. The voxel-based branch learns local neighborhood information at varying receptive field sizes using convolutional layers, and the point-based branch learns high-resolution information that may be lost in the voxelized branch using MLP layers. The information across the two branches are 
fused at intermediate steps to capture complementary information from the input 3D point cloud.

\subsubsection{Point Correspondences}
Given a pair of samples $\{\mathcal{S}_1, \mathcal{S}_2\}$ from nearby locations (i.e. $ \mathcal{D}(\mathtt{x_1},\mathtt{x_2}) < \tau_p $, where $ \tau_p $ is a distance threshold), corresponding points in the two point clouds $\{\mathcal{P}_1, \mathcal{P}_2\}$ are calculated by first transforming the point clouds to a common coordinate frame using the geo-location information $\{\mathtt{x_1},\mathtt{x_2}\}$ followed by ICP~\cite{besl1992method} for better alignment (to account for possible errors in geo-locations). After alignment, point correspondences can be found using radius-based nearest neighbor search. This search is made efficient using the approximate nearest neighbor search algorithm FLANN~\cite{muja2009fast}. For each point $ \tensor[^{1}]{p}{_i}  $ in point cloud $ \mathcal{P}_1 $ and radius $r$, the corresponding points in point cloud $ \mathcal{P}_2 $ are found as $ C^{1 \leftrightarrow 2}_i = \{ (i,j) \, | \, \mathcal{D}(\tensor[^{1}]{p}{_i} ,\tensor[^{2}]{p}{_j}) < r \}$. The set of all corresponding points' indices between the two point clouds is represented as $ C^{1 \leftrightarrow 2} $.

\subsubsection{Local Consistency Loss}
Given two samples $\{\mathcal{S}_1, \mathcal{S}_2\}$, their associated local features  $\{\{f\}_1, \{f\}_2\}$ and the point correspondences $ C^{1 \leftrightarrow 2} $, a contrastive loss is applied to minimize the distance for features of corresponding points (positive pairs) while maximizing the distance of features of non-corresponding points (negative pairs). We adopt the Hardest-Contrastive loss~\cite{Choy_2019_ICCV} which is defined as,
\begin{equation}%
\begin{split}
    \mathcal{L}_{lc} &\!=\!\sum_{(i,j)  \in C^{1 \leftrightarrow 2}}^{} \Bigg\{ \left[||f(\tensor[^{1}]{p}{_i})\!-\!f(\tensor[^{2}]{p}{_j})||_{2}^{2} \!-\! m_p \right ]_{+}  / \,|C^{1 \leftrightarrow 2}| \\ 
    & \!+ \! \, \lambda{_n} I_i \left [ m_n \!-\! \underset{k\in \mathcal{M}}{{\rm min}} ||f(\tensor[^{1}]{p}{_i})\!-\!f(\tensor[^{2}]{p}{_k})||_{2}^{2} \right ]_{+} / \,|C^{1 \leftrightarrow 2}_1| \\
    & \!+ \! \, \lambda{_n} I_j  \left [m_n \!-\! \underset{k\in \mathcal{M}}{{\rm min}} ||f(\tensor[^{2}]{p}{_j})\!-\!f(\tensor[^{1}]{p}{_k})||_{2}^{2} \right ]_{+}  / \,|C^{1 \leftrightarrow 2}_2| \Bigg\},
\end{split}   
    \label{Eq:point_loss}
\end{equation}
where, $\mathcal{M}$ is a  random subset of features used for hard negative mining and  $\left [  . \right ]_{+}$ denotes the hinge loss. $I_i$ is short for $I(i, k_i, r)$, which is an indicator function that returns 1 if the point $k_i$ is non-corresponding (outside radius $r$) to point $i$ and 0 otherwise,
where $k_i = \argmin_{k \in \mathcal{M}} || f_i - f_k||_2$. 
$|C^{1 \leftrightarrow 2}_1| = \sum_{(i,j) \in C^{1 \leftrightarrow 2}} I(i, k_i, r)$ is the total number of valid mined negatives for points in $ \mathcal{P}_1 $ ($| C^{1 \leftrightarrow 2}_2 |$ for $ \mathcal{P}_2 $). 
Hyperparameters $ m_p $, $ m_n $ are scalar margins and $\lambda{_n}$ is a scalar weight.

The local consistency loss $\mathcal{L}_{lc}$ acts on the parameters of $\phi$ in the decomposition of \eqref{eqn:decomposition} to get well-formed repeatable local features. 

\subsection{Global Descriptor}
\label{section:sop}

Given a point cloud $ \mathcal{P} = \{p_i\} $ and the set of point features $ \{f(p_i)\} $ where $ f(p_i) \in \mathbb{R}^{d} $, the second-order pooling $ F^{O_2} $ of the features is defined as,
\begin{equation}\label{eqn:o2p}
 F^{O_2} = \{F^{O_2}_{xy}\} , \quad F^{O_2}_{xy} = \max_{p_i \in \mathcal{P} } \ f_{xy}^{o_2}(p_i), 
\end{equation}
where $ F^{O_2} $ is a matrix with elements $ F^{O_2}_{xy} ( 1 \leq x,y \leq d) $ 
and $ f^{o_2}(p_i) = f(p_i)f(p_i)^T \in \mathbb{R}^{d \times d} $ is the outer product of the point feature with itself. This accounts to taking the element-wise maximum of the second-order features of all points in the point cloud. 

In order to make the scene descriptor matrix $ F^{O_2} $ more discriminative, we use Eigen-value Power Normalization (ePN) \cite{9521687, 7439823, Li2017}. Given the singular value decomposition $ F^{O_2} = U\lambda V^T $ the ePN result $ F^{O_2}_{\alpha} $ is obtained by raising each of the singular values by a power of $ \alpha $ as follows,
\begin{equation}\label{eqn:o2p_nlt}
 F^{O_2}_{\alpha} = U\hat{\lambda}V^T, \quad  \hat{\lambda} = \textit{diag}(\lambda^{\alpha}_{1,1},  .. , \lambda^{\alpha}_{d,d}),
\end{equation}
where, $ \alpha = 0.5 $. The matrix $ F^{O_2}_{\alpha} $ is flattened and normalized to obtain the final global descriptor vector $g \in \mathbb{R}^{d^2} $. To incorporate \eqref{eqn:o2p_nlt} into our end-to-end pipeline, we utilize differentiable SVD as introduced in \cite{Papadopoulo2000diffsvd}, and its PyTorch implementation.

After the aggregation of point features into a global descriptor using second-order pooling, the scene-level loss $ \mathcal{L}_g $ is applied to a tuple $ \mathcal{T} $ of training samples. We use the quadruplet loss~\cite{chen2017beyond} where a tuple is denoted as $\mathcal{T}_i = (\mathcal{S}_{a}, \{\mathcal{S}_{p}\}, \{\mathcal{S}_{n}\}, \mathcal{S}_{n^{*}} )$, where $\mathcal{S}_{a}$ is the anchor sample, \{$\mathcal{S}_{p}$\} a set of positives (such that $ \mathcal{D}(\mathtt{x_a},\mathtt{x_p}) < \tau_p $), \{$\mathcal{S}_{n}$\} a set of negatives (such that $ \mathcal{D}(\mathtt{x_a},\mathtt{x_n}) > \tau_n $), and $ \mathcal{S}_{n^{*}} $ is sampled such that it is not a positive to the query nor to all previous negatives (such that $ \mathcal{D}(\mathtt{x_{n^{*}}},\mathtt{x}) > \tau_p \, \forall \, \mathtt{x}=\{\mathtt{x_a}, \mathtt{x_n}\} $).

In each tuple we first find the hardest positive sample,
 \begin{equation}
    \mathcal{P}_{hp}=\underset{\mathcal{P}_{p^{i}}\in\left \{ \mathcal{P}_{p} \right \}}{ \rm max}\, \, ||g(\mathcal{P}_{a})-g(\mathcal{P}_{p^{i}})||_{2},
    \label{Eq:HP}
\end{equation}
The quadruplet loss is then defined as:

\begin{equation}
\begin{split}
    \mathcal{L}_g &\!= \sum_{i}^{\mathcal{N}} \Bigg\{ \left[||g(\mathcal{P}_{a})\!-\!g(\mathcal{P}_{hp})||_{2}^{2}\!-\!||g(\mathcal{P}_{a})\!-\!g(\mathcal{P}_{n^{i}})||_{2}^{2}\!+\!\alpha \right ]_{+} \\ 
    & \!+ \!  \left [||g(\mathcal{P}_{a})\!-\!g(\mathcal{P}_{hp})||_{2}^{2}\!-\!||g(\mathcal{P}_{n^{*}})\!-\!g(\mathcal{P}_{n^{i}})||_{2}^{2}\!+\!\beta \right ]_{+} \Bigg\},
\end{split}   
    \label{Eq:quadruplet}
\end{equation}
where, $\alpha$ and $\beta$ are constant margins and $\mathcal{N}$ is the number of sampled negatives.

\subsection{Joint Local and Global Loss}
\label{section:loss}

Our network is jointly optimized by a weighted sum of the global scene-level loss and the local consistency loss described as: 

\begin{equation}
\label{eq:final_loss}
\mathcal{L} = \mathcal{L}_g + \omega \cdot \mathcal{L}_{lc}
\end{equation}
where, $\omega$ is a scalar hyperparameter term.  
\section{Experimental Setup}
\label{sec:experiments}

\subsection{Implementation and Training Setup}

The proposed network is implemented using the PyTorch framework and trained on 12 Nvidia Tesla P100-16GB GPUs using $ \mathtt{DistributedDataParallel}$. The TorchSparse library~\cite{tang2020searching} is used for sparse convolutions. 
During training, the ground plane is first removed using RANSAC plane fitting followed by down-sampling using a voxel grid filter of $10cm$. Finally, input point clouds are limited to a maximum of $35K$ points. 
To reduce overfitting, we apply the following data augmentations for training. Random point jitter is introduced using Gaussian noise sampled from $\mathcal{N}(\mu=0, \sigma=0.01)$ clipped at $0.03m$. Each point cloud is also randomly rotated about the $z$-axis by an angle between $ \pm 180^{\circ}$. Note that ground plane removal is not used during evaluation to speed up inference time as it does not affect evaluation performance of our proposed method\footnote{The additional features from the ground plane are still well-formed (property in Fig. \ref{fig:hero_image}), and after aggregation, the global descriptors still remain discriminative due to the robustness of ePN \eqref{eqn:o2p_nlt} to feature burstiness \cite{9521687}.}.

For a fair comparison with PointNetVLAD~\cite{Uy2018}, we set the same global descriptor dimension ($ d^2 = 256 $), hence the dimension of the local features is set to $ d = 16 $. 
In the local consistency loss $ \mathcal{L}_{lc}$, the margins $m_p$ and $m_n$ are set to 0.1 and 2.0 respectively, and $\lambda{_n}$ is set to 0.5. 
The quadruplet loss margins are set to $\alpha=0.5$ and $\beta=0.3$. The distances for sampling positive and negative point cloud pairs are set to $\tau_p=3 m$ and $\tau_n=20 m$.
For $ \mathcal{L}_g$ we use 2 positives, 9 negatives and 1 other negative. 
We train our model using the Adam optimizer with an initial learning rate of 0.001 and a multi-step scheduler to drop the learning rate by a factor of 10 after 10 epochs and train until convergence for a maximum of 24 hours. 

\subsection{Datasets}

We evaluate the proposed method on two public LiDAR datasets (KITTI, MulRan), both of which were collected from a moving vehicle in multiple dynamic urban environments. Note that these datasets are collected using different LiDAR sensors (Velodyne HDL-64E in KITTI, Ouster OS1-64 in MulRan) and in different countries (Germany, Korea). 

{\parskip=5pt
\noindent\textit{KITTI}: The KITTI odometry dataset~\cite{Geiger2013} contains 11 sequences of Velodyne  HDL-64E LiDAR scans collected  in Karlsruhe, Germany. We train on these 11 sequences using the leave-one-out cross-validation strategy and evaluate on the 6 sequences with revisits (00, 02, 05, 06, 07 and 08).
}

{\parskip=5pt
\noindent\textit{MulRan}: The MulRan dataset~\cite{mulran20} contains scans collected from an Ouster-64 sensor from multiple environments in South Korea. The dataset contains 12 sequences, 9 of which we use for evaluation.
We train on DCC1, DCC2, Riverside1 and Riverside3 sequences and evaluate on the remaining sequences of DCC, Riverside and KAIST. 
To assess the generalization capabilities of methods, the KAIST sequences are unseen test sets for evaluation. 

}

\subsection{Evaluation Criteria}
We compute the cosine similarity between the global descriptors of each query with a database of global descriptors of previously seen point clouds in each sequence. Previous entries adjacent to the query by less than $ t_r $ time difference are excluded from the search to avoid matching to the same instance. For $ t_r $ we use $30 s$ and $90 s$ for KITTI and MulRan, respectively. 
Methods are compared using the Precision-Recall curve and its scalar metric the maximum $ F1 $ score ($ F1_{max} $). The 3m, 20m thresholds are used to classify true positives and false positives respectively, as done in \cite{vidanapathirana2020locus}.

\section{Results}
\label{sec:results}

We first demonstrate the performance improvement from the inclusion of $\mathcal{L}_{lc}$. We evaluate our method in comparison to other state-of-the-art methods and conduct a run-time analysis to judge the suitability for real-time operation.  

\subsection{Ablation study on point-wise loss}
\begin{table}[t]
\centering
\begin{tabular}{c c c c} 
 \hline
 Method & DCC2 & Riverside2 & mean  \\ [0.5ex] 
 \hline
  $\mathcal{L}_g$    & 0.355 & 0.472 & 0.413  \\
  $\mathcal{L}_g + 0.1\cdot\mathcal{L}_{lc}$   & 0.471 & 0.578 & 0.524  \\
 \textbf{ $\mathcal{L}_g + 1.0\cdot\mathcal{L}_{lc}$}    & \textbf{0.591} & \textbf{0.747} & \textbf{0.669} \\
 \hline
\end{tabular}
\caption{Ablation study on the effects of the $\omega$ term for our joint loss. ${L}_g$ is the scene-level loss and ${L}_{lc}$ is the local consistency loss.}
\label{table_ablation}
\end{table}
We evaluate the effect of inclusion of the local consistency loss through an ablation study on selected sequences of the MulRan dataset. We train on sequences DCC1, Riverside1 and evaluate performance on DCC2 and Riverside2.
\tabref{table_ablation} summarizes the $F1_{max}$ for each test sequence by varying the weight of the local consistency loss  $\omega$ in~\eqref{eq:final_loss}. It is evident that the inclusion on the local consistency loss leads to an improvement in place recognition performance. $\omega = 0.1$ leads to an improvement of $26\%$ mean $F1_{max}$ with respect to the baseline while $\omega = 1.0$ leads to an improvement of $61\%$ leading to the best performance. All the following experiments are carried out with $\omega = 1.0$.

\begin{figure}[t!]
\centering
\includegraphics[width=0.48\textwidth]{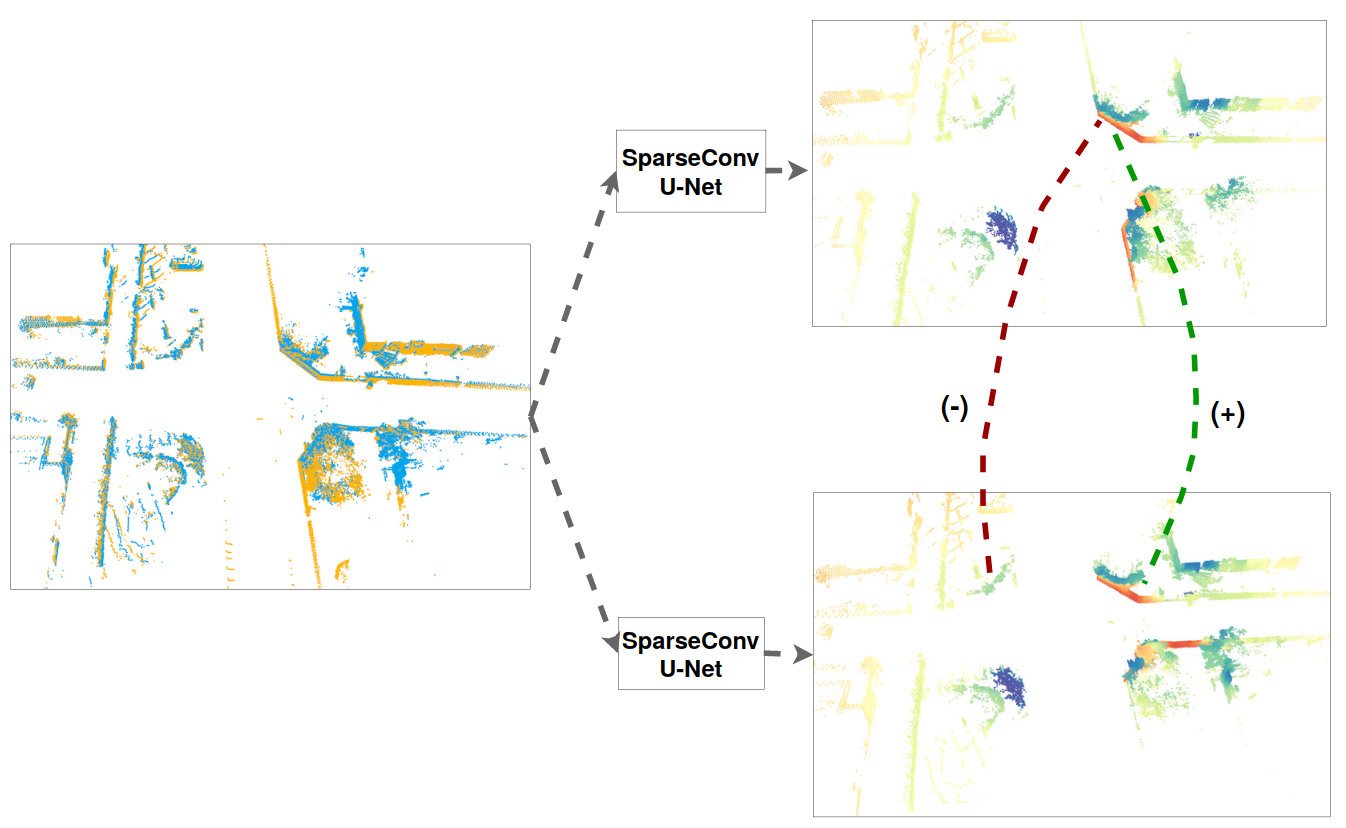}
\caption{t-SNE of point-features extracted from two point clouds of the same location shows that our method produces `locally consistent' features. 
Colors represent the similarity of point features of one cloud relative to the other.}
\label{fig:hero_image}
\end{figure}

A qualitative depiction of the effect of $\mathcal{L}_{lc}$ is shown in Fig. \ref{fig:hero_image} between two point clouds extracted from nearby locations. The left half shows the alignment of the point cloud after ground plane removal to estimate point correspondences during training. The right half shows the two point clouds separately with each point colored based on the t-SNE embedding of the local features extracted using our pre-trained model (note that point correspondence information is not used during inference). The visualization clearly highlights that distinct regions in a single point cloud are in distinct regions in the feature space. Additionally, across point clouds, corresponding regions have similar point features implying that the local features extracted from our method are repeatable.

\subsection{Comparison to State-of-the-Art}
\begin{table*}[htbp]
\begin{center}
\begin{tabular}{l@{\quad}|r@{\quad}r@{\quad}r@{\quad}r@{\quad}r@{\quad}r@{\quad}|r@{\quad}|r@{\quad}r@{\quad}r@{\quad}r@{\quad}r@{\quad}|r@{\quad}|r@{\quad}r|}
\cline{2-14}
&  \multicolumn{7}{c|}{KITTI} & \multicolumn{6}{c|}{MulRan} \\
\cline{2-14}
&  \begin{tabular}{@{}c@{}}00\end{tabular}
& \begin{tabular}{@{}c@{}}02\end{tabular}
& \begin{tabular}{@{}c@{}}05\end{tabular}
& \begin{tabular}{@{}c@{}}06\end{tabular}
& \begin{tabular}{@{}c@{}}07\end{tabular}
& \begin{tabular}{@{}c@{}}08\end{tabular}
& \begin{tabular}{@{}c@{}}mean\end{tabular}
& \begin{tabular}{@{}c@{}}K1\end{tabular}
& \begin{tabular}{@{}c@{}}K2\end{tabular}
& \begin{tabular}{@{}c@{}}K3\end{tabular}
& \begin{tabular}{@{}c@{}}D3\end{tabular}
& \begin{tabular}{@{}c@{}}R2\end{tabular}
& \begin{tabular}{@{}c@{}}mean\end{tabular}
\\[2pt]
\hline
ScanContext~\cite{Kim2018} & 0.966 & 0.871 & 0.914 & 0.985 & 0.698 & 0.610 & 0.841 & 0.954 & \textbf{0.969} & \textbf{0.994} & 0.893  & 0.826 & 0.916 \\
PointNetVLAD~\cite{Uy2018} & 0.909 & 0.637 & 0.859 & 0.924 & 0.171 & 0.437 & 0.656 & 0.952 & 0.856 & 0.979 & 0.685 & 0.868 & 0.868 \\
Locus~\cite{vidanapathirana2020locus} & \textbf{0.983} & 0.762 & \textbf{0.981} & \textbf{0.992} & \textbf{1.000} & \textbf{0.931} & \textbf{0.942} & 0.938 & 0.874 & 0.969 & 0.718 & \textbf{0.994} & 0.899 \\ 
\textbf{LoGG3D-Net (Ours)}  & 0.953 & \textbf{0.888} & 0.976 & 0.977 & \textbf{1.000} & 0.843 & 0.939 & \textbf{0.966} & 0.938 & 0.991 & \textbf{0.977} & 0.969 & \textbf{0.968} \\
\hline
\end{tabular}
\end{center}
\caption{Evaluation of sequential place recognition on the KITTI and MulRan datasets using the $F1_{max}$ metric under the $3m,20m$ revisit criteria.
}
\label{tab:seq_eval}
\end{table*}

We compare \textit{LoGG3D-Net} with the state-of-the-art handcrafted method ScanContext\footnote{\url{https://github.com/irapkaist/scancontext}}~\cite{Kim2018}, the recently proposed hybrid method Locus\footnote{\url{https://github.com/csiro-robotics/locus}}~\cite{vidanapathirana2020locus}, and the popular end-to-end method PointNetVLAD~\cite{Uy2018}. For ScanContext we use the python version of the code provided by the authors. We use the 20x60 descriptor size and use ring-key search to find the top-10 candidates for descriptor matching. For PointNetVLAD we use a PyTorch-based re-implementation of the original tensorflow implementation\footnote{\url{https://github.com/mikacuy/pointnetvlad}}. 

The results are summarized in~\tabref{tab:seq_eval}. On the KITTI dataset, Locus remains the highest performing method with a mean $F1_{max}$ score $1\%$ higher than ours. On the MulRan dataset we obtain the best mean  $F1_{max}$ score which is $5\%$ higher than the next best performing method,~\ie{} ScanContext. It should be noted that the local feature extractor of Locus was trained on the KITTI sequences 05 and 06 thus leading to its extremely high performance on KITTI and relatively low performance on MulRan. 

\begin{figure}[!tbp]
  \centering
  \subfloat[KITTI 02.]{\includegraphics[width=0.23\textwidth]{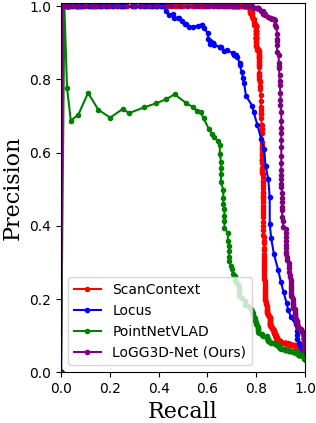}\label{fig:kitti02}}
  \hfill%
  \subfloat[MulRan DCC 03.]{\includegraphics[width=0.23\textwidth]{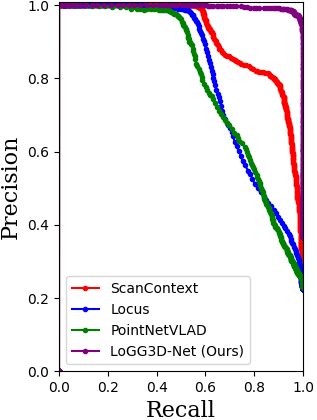}\label{fig:mulrandcc3}}
  \caption{Precision-Recall curves for evaluation on select sequences of the KITTI and MulRan datasets. }
  \label{fig:pr_curve}
\end{figure}

Precision-Recall plots for the sequences KITTI 02 and DCC 03 are depicted in~\figref{fig:pr_curve}. KITTI 02 contains several repetitive environments and a single revisit to an intersection in the opposite direction. DCC 03 contains long traversals of revisits from the reverse direction. All these scenarios prove challenging for most methods while the adverse effect on the performance of \textit{LoGG3D-Net} is not as pronounced.

The feature dimensions of Locus and ScanContext give these methods an unfair advantage in terms of representation power. The ScanContext descriptor is 20$\times$60 (a total of 1200 floating point numbers) and the Locus descriptor is 4096 dimensions. Both \textit{LoGG3D-Net} and PointNetVLAD have descriptor sizes of 256 which are compact and scale well to large databases essential for real-time robotic applications. 

\subsection{Runtime Analysis}
The computation time for pre-processing, description and querying is demonstrated in ~\tabref{tab:runtime}.
All experiments in this section are run on a system with an 8 core Intel i7-9700 processor with 32GB RAM and a single Nvidia RTX 2080Ti GPU.
Since the time for retrieval increases with the size of the database, we present the average retrieval time for all methods on MulRan DCC1 which consists of 5541 point clouds. ScanContext uses ring-key retrieval to find the top-10 candidates for full descriptor distance calculation.  

The results show that Locus has very high pre-processing time due to ground-plane removal and a high description time (due to around 30-80 sequential forward passes through the segment feature extraction network which is not parallelized). ScanContext has very high retrieval time even after limiting the number of ring-key candidates to just 10. The need for cosine similarity computation on each column shifted variant of the descriptor increases the descriptors query time.  PointNetVLAD is the most efficient method for descriptor extraction due to its light network architecture. However, we note that PointNetVLAD uses a considerable amount of pre-processing to first remove the ground plane and then iteratively downsample the point cloud to exactly 4096 points which makes its total time longer than ours. 

Our method has the lowest pre-processing and retrieval times allowing it to run at real-time ($\sim{10Hz}$), enabling its integration into SLAM systems as a loop closure detection module. 

\begin{table}[H]
  \centering
\begin{tabular}{l|ccc|c}

   & Pre- & Description & Querying & Total  \\
   & proc. &  &  &   \\
	\hline 
	 ScanContext~\cite{Kim2018} & 63 & 582 & 3968 & 4613 \\
	 PointNetVLAD~\cite{Uy2018} & 534 & 7 & 1 & 542 \\
	 Locus~\cite{vidanapathirana2020locus} & 573 & 633 & 6 & 1212 \\
	 \textbf{LoGG3D-Net (Ours)} & 15 & 74 & 1 & \textbf{90} \\
	\hline
\end{tabular}
\caption{Runtime analysis: Average time taken on MulRan DCC1  (in ms).
  }
  \label{tab:runtime}
\end{table}

\section{Conclusion}
\label{sec:conclusion}
This paper introduced the use of a local consistency loss in addition to a global contrastive loss for the training of end-to-end models for 3D place recognition. This additional constraint enforces corresponding points in different point clouds of the same place to have similar embeddings. 
The implementation, named \textit{LoGG3D-Net}, is based on a U-Net architecture which uses sparse point-voxel convolution to enable efficient and fine-grained inference on high resolution point clouds. Second-order pooling along with differentiable Eigen-value power normalization ensures that point clouds are encoded into a single vector representation which better captures the distribution of local features. Evaluation of \textit{LoGG3D-Net} on 11 sequences of two large-scale public benchmarks (KITTI and MulRan) resulted in mean $F1_{max}$ scores of $0.939$ and $0.968$ on KITTI and MulRan respectively, achieving state-of-the-art performance. 
Ablation studies demonstrated that the local consistency loss provides a consistent and significant improvement. Run-time analysis demonstrated real-time inference enabling integration
into SLAM systems as a loop closure detection module.

\balance{}

\bibliographystyle{./bibliography/IEEEtran}
\bibliography{main}

\end{document}